\def\year{2017}\relax
\documentclass[letterpaper]{article}
\usepackage{aaai17}
\usepackage{times}
\usepackage{helvet}
\usepackage{courier}
\usepackage{hyperref}       
\usepackage{url}            
\usepackage{booktabs}       
\usepackage{amsfonts}       
\usepackage{nicefrac}       
\usepackage{microtype}      
\usepackage{color}
\usepackage{graphicx}
\usepackage{amsmath}
\usepackage{bm}
\usepackage{multirow}
\usepackage{subfigure}
\usepackage{soul}
\usepackage{lipsum,multicol}

\usepackage{tabularx}
\usepackage{arydshln}
\newcolumntype{C}[1]{>{\centering\let\newline\\\arraybackslash\hspace{0pt}}p{#1}}
\newcolumntype{L}[1]{>{\raggedright\let\newline\\\arraybackslash\hspace{3pt}}p{#1}}

\newcommand{\etal}{\textit{et al}.}

\begin{document}
%
\title{Training Recurrent Answering Units  with Joint Loss Minimization for VQA}
\author{Hyeonwoo Noh \hspace{1cm} Bohyung Han \\
Dept. of Computer Science and Engineering, POSTECH, Pohang, Korea \\
\texttt{\{shgusdngogo,bhhan\}@postech.ac.kr}\\
}
\maketitle
\begin{abstract}
We propose a novel algorithm for visual question answering based on a recurrent deep neural network, where every module in the network corresponds to a complete answering unit with attention mechanism by itself.
The network is optimized by minimizing loss aggregated from all the units, which share model parameters while receiving different information to compute attention probability.
For training, our model attends to a region within image feature map, updates its memory based on the question and attended image feature, and answers the question based on its memory state.
This procedure is performed to compute loss in each step.
The motivation of this approach is our observation that multi-step inferences are often required to answer questions while each problem may have a unique desirable number of steps, which is difficult to identify in practice.
Hence, we always make the first unit in the network solve problems, but allow it to learn the knowledge from the rest of units by backpropagation unless it degrades the model.
To implement this idea, we early-stop training each unit as soon as it starts to overfit.
Note that, since more complex models tend to overfit on easier questions quickly, the last answering unit in the unfolded recurrent neural network is typically killed first while the first one remains last.
We make a single-step prediction for a new question using the shared model.
This strategy works better than the other options within our framework since the selected model is trained effectively from all units without overfitting.
The proposed algorithm outperforms other multi-step attention based approaches using a single step prediction in VQA dataset.
\end{abstract}


\section{Introduction}
\label{sec:intro}

Visual Question Answering (VQA)~\cite{VQA} is the problem to answer questions related to an input image.
Contrary to traditional visual recognition tasks, which concentrate on a predefined problem such as object detection, scene classification, activity recognition, etc., VQA involves various recognition tasks at the same time and provides a unified approach to solve the problems defined by questions.
Due to these reasons, VQA requires substantial amount of learning to capture information from images and understand questions.

Recently, deep learning based approaches using multiple reasoning steps with attention~\cite{SAN,DMNQA} have been proposed to improve the performance of VQA systems.
After extracting features from an image and a question using Convolutional Neural Network (CNN) and Recurrent Neural Network (RNN), respectively, these methods iteratively generate answers by combining question feature with image feature of local area given by attention.
This approach is natural as many questions in VQA conceptually require multiple steps for reasoning.
For example, to answer the question like ``what is in front of the giraffe?'' the VQA model should be able to solve the subtasks in sequel as ``finding a giraffe,'' ``identifying the region in front of the giraffe'' and ``classifying object in the specified region.''

One of the critical limitations of the existing methods is that the number of steps for reasoning is fixed to a predefined number.
Even though the exact number of steps required to answer a question is difficult to figure out, it is reasonable to assume that different questions need the different numbers of steps to reach solutions.
For example, a question ``what color is the apple on the table?'' requires more steps than another question ``what color is the apple?''
If the number of inference steps is not adaptive to questions, we may need to solve multiple subtasks in a single step or a single subtask in multiple steps, which are likely to degrade system performance by causing overfitting or underfitting of trained models.

Instead of presetting the number of steps, we make the model learn the number of steps for inference of each question implicitly by simply minimizing joint loss from multiple steps of our recurrent deep neural network without extra supervision.
We believe that learning to predict with a proper number of steps not only enhance the overall accuracy but also help each answering unit to generalize better.
However, identifying the optimal number of steps is an extremely difficult task, and we propose an indirect but simple solution to learn a unified model for inference.
By analyzing the answers from each step, we find out that predictions with more steps tend to overfit to easier questions even if later steps could solve more complex questions better than the earlier ones.
Based on this observation, we progressively stop backpropagating the loss from the overfitting units.
Training is terminated when the accuracy of the first answering unit is saturated.
With this training strategy, we empirically show that the single-step prediction based on the first unit, whose model is trained with joint loss from multiple steps, outperforms other training-testing variations, and the accuracy is further improved by integrating early-stopping training strategy.
%
The main contribution of the proposed algorithm is summarized below:
\begin{itemize}
\item We propose a novel architecture based on a deep recurrent neural network for VQA, which is composed of multiple answering units with shared parameters and optimized by minimizing joint loss from multiple units.
\item We show that the VQA model involving multiple reasoning steps tends to overfit to easier questions quickly, and develop a unique training strategy to early-stop overfitting units progressively until the accuracy of the first unit is saturated.
\item The proposed algorithm outperforms other multi-step attention based approaches using a single step prediction in VQA dataset.
\end{itemize}

The rest of the paper is organized as follows. 
We first review related work in Section~\ref{sec:related} and discuss our main idea with motivation in Section~\ref{sec:overview}.
Section~\ref{sec:algorithm} describes the architecture to implement our idea and the training and testing methods of the proposed algorithm.
We present experimental results of our algorithm in the standard public benchmark dataset in Section~\ref{sec:experiment}.


\section{Related Work}
\label{sec:related}

The VQA problem is first addressed by \cite{Multiworld}, which proposes a Bayesian framework to combine symbolic reasoning and uncertain visual information.
\cite{zhou2015simple} has proposed a shallow neural network for VQA, which accepts the CNN features of images and the Bag-of-Words (BoW) representations of questions as its inputs.
Recently, VQA algorithms are often formulated with deep neural networks since the techniques based on deep learning have straightforward end-to-end training procedure by  backpropagation and present competitive performance in terms of accuracy.
These approaches typically pose VQA problems as simple classification tasks based on the joint features from images and questions, where CNNs and LSTMs~\cite{LSTM} are employed to encode images and questions, respectively \cite{mren2015,Askneurons} while only CNNs are used for both encoding and classification in \cite{Convqa}.

Several VQA systems~\cite{SAN,DMNQA,Wheretolook} attempt to identify relevant regions in the input image to answer questions, which is often performed by the soft attention mechanism~\cite{Showattend}.  
Specifically, \cite{Wheretolook} utilizes a single step attention to an object proposal for VQA while \cite{DMNQA} also employs multi-step attention based on the dynamic attention network~\cite{DMN}. 
Stacked attention network~\cite{SAN} is motivated by memory networks~\cite{Memnet,DMN}, and constructs two succesive attention modules for multi-step inferences.

Several approaches employ the network architectures adaptive to input questions.
Dynamic parameter prediction network~\cite{DPPnet} determines the parameters of a fully connected layer in CNN given a question, and constructs a unified model to handle various tasks related to input images.
With similar motivation, neural module networks combine multiple module networks to construct a single deep network~\cite{NMN,NMN2}, where the module combinations are given by syntactic parsing of questions. 

Deep neural networks are sometimes trained with multiple supervisions to facilitate training procedure, where losses are backpropagated from multiple branches.
GoogLeNet~\cite{Googlenet} attaches auxiliary classifiers to a few intermediate layers of the network to provide additional supervision for training.
Deeply supervised nets~\cite{DSN} has a companion objective function to avoid the vanishing gradient problem. 
Deeply recursive convolutional network~\cite{kim2016deeply} provides supervision to every recurrent convolutional layer and makes inference based on an ensemble of individual predictions.
Our approach has something common with \cite{kim2016deeply} in that supervision is given to each recurrent unit, but is differentiated since we analyze role of each supervision and propose a novel training strategy of early stopping.

\section{Overview}
\label{sec:overview}
This section describes the general formulation of VQA and discusses our approach based on recurrent deep neural network with its motivation.

\subsection{Problem Formulation}
We formulate VQA problem as a classification task. 
For a given image $I$ and a question $q$, a VQA model predicts the best answer $a^*$, which is formally given by
\begin{equation}
a^* = \underset{a\in{\Omega}}{\operatorname{argmax}} ~ p({a} \vert {I}, q;{\bm{\theta}})
\label{eq:objective}
\end{equation}
where $\Omega$ denotes a set of all possible answers and ${\bm{\theta}}$ is a set of model parameters in the network. 
There exist a lot of different methods to implement this probabilistic framework, but we focus on the models based on deep neural networks, which are frequently employed for VQA problems in recent years.

The models based on deep neural networks are typically composed of three main components: image encoder, question encoder, and answering module. 
Image and question encoders extract image feature ${\bf f}_I$ from an input image $I$ and question feature ${\bf f}_q$ from a question sentence $q$, respectively. 
Answering module takes extracted image and question features, and generates an answer by combining information from the image and the question.
Therefore, the module should be able to perform various tasks defined by questions.
In this work, we propose a deep neural network architecture and investigate a training strategy for the answering module.

\subsection{Main Idea}
The motivation behind this work is our observation that solving a task defined by a question often requires the capability to solve a sequence of atomic subtasks.
However, the kinds and numbers of the subtasks in the sequence vary in individual questions, and, in addition, the same subtasks may appear in any places within the sequence depending on the question.
Due to these reasons, it is extremely difficult to develop a unified algorithm to handle all the variations.
Instead, to overcome the challenges indirectly, we design a novel neural network architecture for VQA and propose a training strategy with early-stopping based on a simple joint loss minimization.

\begin{figure*}[t]
\centering
\includegraphics[width=0.95\linewidth] {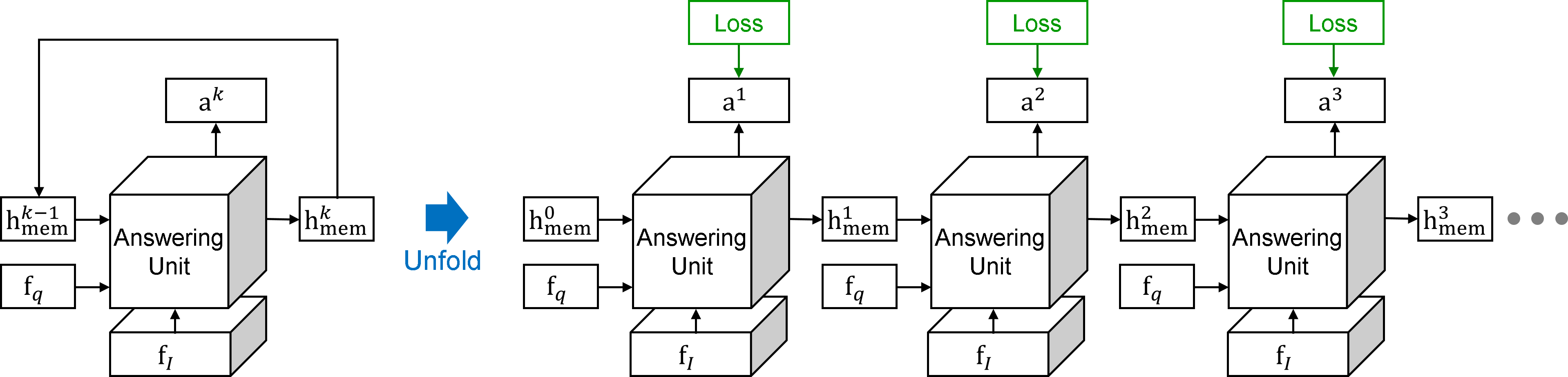}
\caption{
Overall architecture of the proposed network.
The proposed network is a recurrent deep neural network, where each recurrent unit corresponds to a complete module for visual question answering.
For training, we unfold the network to predict answer and give supervision for every steps.
For testing, we use a single answering unit to answer a question about an image.
}
\label{fig:overall_network}
\end{figure*}

The overall architecture of the proposed algorithm is illustrated in Figure~\ref{fig:overall_network}.
The main component of the proposed network is an answering unit. 
Each answering unit is capable of solving the full task; based on the features extracted from image and question, it predicts an answer and updates its hidden state.
By concatenating multiple answering units sequentially, we infer a series of answers, which are predicted by the model integrating subtasks progressively and solve more and more composite tasks.
This procedure is implemented as a recurrent neural network whose recurrent unit corresponds to each answering unit.
%

We need to train the proposed network so that it can answer a given question by solving a series of subtasks one by one using the answering units.
The main challenge to this objective is that it is difficult to know the optimal number of steps to solve each problem since there are various question types with different complexity.
To circumvent this problem, we always make the first unit in the network solve problems, but allow it to learn the knowledge from the rest of units by backpropagation unless it degrades the model.
For the purpose, we simply provide the same supervision to every step in the unfolded recurrent neural network and optimize all the answering units with shared parameters jointly using the standard backpropagation technique for RNNs.
We also propose a unique training technique with early-stopping strategy, which is useful to avoid overfitting in complex models among multiple answering units.


\section{Algorithm}
\label{sec:algorithm}
This section discusses the proposed answering module and the procedure of training and testing.
We provide more detailed description about the answering modules in the supplementary document.

\subsection{Answering Module}

Answering module is a recurrent neural network, which is composed of multiple answering units with shared model parameters as illustrated in Figure~\ref{fig:overall_network}.
Using a given image feature map ${\bf f}_I \in \mathbb{R}^{P \times L}$ (with $P$ channels and $L$ locations), a question feature ${\bf f}_q \in \mathbb{R}^{Q}$ and an initial memory state ${\bf h}^0_\mathrm{mem} \in \mathbb{R}^{R}$, answering module provides a sequence of answer probability vectors ${\bf a}^k \in \mathbb{R}^C$ ($k = 1, \dots, K$), where $K$ is the number of answering units.
Formally, the $k^{\rm th}$ answering unit predicts an answer probability as
\begin{equation}
{\bf a}^k               =       \mathrm{Answer}_k      (      {\bf f}_I       ,      {\bf f}_q       ,      {\bf h}^{k-1}_\mathrm{mem}     ),
\label{eq:ans_operation}
\end{equation}
where $\mathrm{Answer}_k(\cdot)$ is the answering operation in the $k^{\rm th}$ answering unit.

\begin{figure}[t]
    \centering
    \includegraphics[width=1\linewidth] {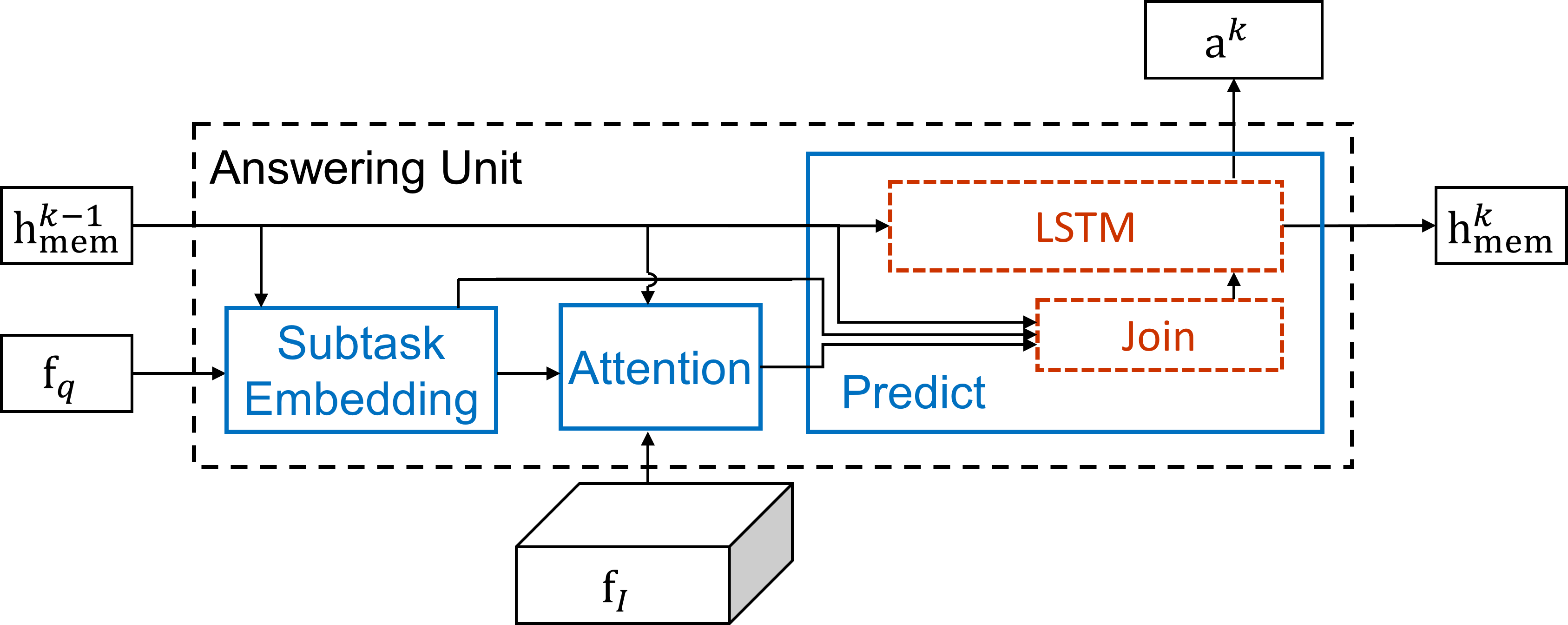}
    \caption{
        The illustration of the answering unit which comprises subtask embedding, attention and predict operation.
    }
    \label{fig:answering_unit}
\end{figure}


\begin{figure}[t]
    \centering
    \includegraphics[width=0.7\linewidth] {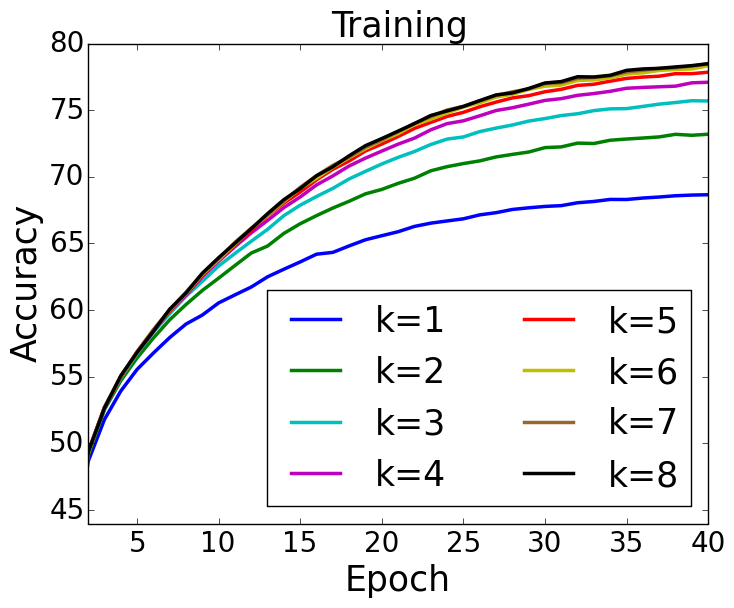} \\
    \centerline{(a) Training accuracy} \vspace{0.2cm}
    \includegraphics[width=0.7\linewidth] {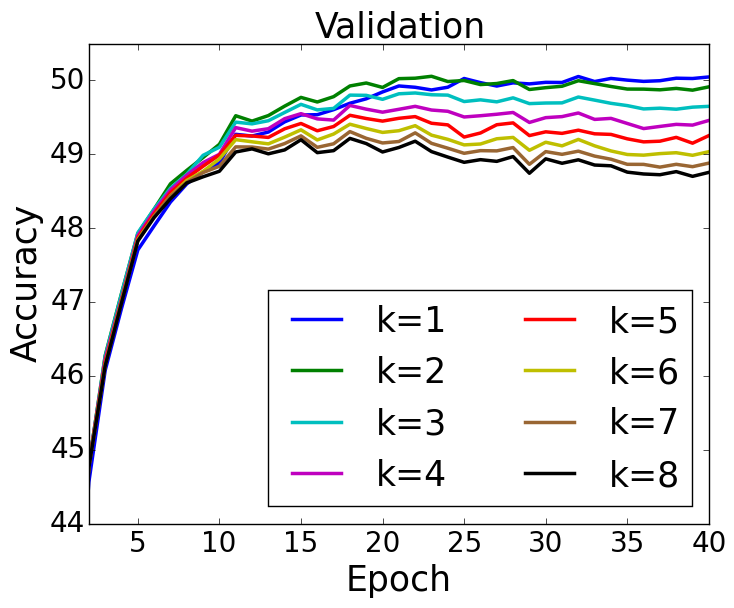} \\
    \centerline{(b) Validation accuracy}
    \caption{
        Training and validation accuracy curve for varying $k$. 
        The prediction from the later step shows higher training accuracy but lower validation accuracy.
    }
    \label{fig:multistep_overfit}
\end{figure}

The function $\mathrm{Answer}_k(\cdot)$ requires several internal operations to predict answer probability as illustrated in Figure~\ref{fig:answering_unit}.
The first operation is to compute a subtask feature, which is extracted based on the question and the history of performed subtasks in the previous steps as follows:
\begin{equation}
{\bf f}_\mathrm{sub}^k = \mathrm{Subtask}_k (    {\bf f}_q    ,    {\bf h}^{k-1}_\mathrm{mem} ; {\bm \theta}_\mathrm{sub}  ) ,
\end{equation}
where ${\bm \theta}_\mathrm{sub}$ is parameter for the subtask module.

This subtask feature is used to perform an attention operation, which finds a relevant location in an image feature map with its corresponding feature vector based on the key ${\bf f}_\mathrm{sub}^k$.
We employ soft attention mechanism~\cite{Showattend} and the attention operation is formally given by
\begin {equation}
\left(  {\bf f}_\mathrm{loc}^k , {\bf f}_\mathrm{att}^k  \right) =  
          \mathrm{Attend}_k  (  {\bf f}_I  , {\bf f}_\mathrm{sub}^k , {\bf h}^{k-1}_\mathrm{mem} ; {\bm \theta}_\mathrm{att} ),
\end {equation}
where ${\bm \theta}_\mathrm{att}$ is the parameters for soft attention, and ${\bf f}^k_\mathrm{loc}$ and ${\bf f}^k_\mathrm{att}$ denote attention probability map and attended feature, respectively.

The last operation is $\mathrm{Predict}_k (\cdot)$ function.
It receives subtask feature, attention information, and the memory state in the previous step to produce the final answer and updates hidden state.
This operation involves LSTM to update hidden state in memory and the answer probability is given by applying a softmax function.
The prediction operation is formally defined as
\begin {equation}
\left( {\bf a}^k , {\bf h}^k_\mathrm{mem} \right) =
    \mathrm{Predict}_k  (  {\bf f}_\mathrm{sub}^k ,
                                        {\bf f}_\mathrm{loc}^k ,
                                        {\bf f}_\mathrm{att}^k ,
                                        {\bf h}^{k-1}_\mathrm{mem} ),
\label{eq:pred_operation}
\end {equation} 
and the two outputs are specifically given by
\begin{align}
{\bf a}^k &= \mathrm{softmax} ( {\bf f}_\mathrm{sub}^k ,
                                               {\bf f}_\mathrm{loc}^k , 
                                               {\bf f}_\mathrm{att}^k ,
                                               {\bf h}_\mathrm{mem}^k  ; {\bm \theta}_\mathrm{soft} ) \\
{\bf h}^k_\mathrm{mem} &= \mathrm{LSTM} ( \{ {\bf f}_\mathrm{sub}^k ,  
                              {\bf f}_\mathrm{loc}^k , 
                              {\bf f}_\mathrm{att}^k \} , 
                              {\bf h}^{k-1}_\mathrm{mem} ; {\bm \theta}_\mathrm{LSTM} ),
\end{align}
where ${\bm \theta}_\mathrm{LSTM}$ is the parameters for LSTM and ${\bf \theta}_\mathrm{soft}$ is the parameters for softmax classifier.

To implement the Answering module, we need to learn the parameters for the all internal operations, and it is performed by the standard backpropagation technique.
The detailed training procedure is described next.

\subsection{Training}

We minimize joint loss from multiple steps by providing the ground-truth answer ${\bf y} \in \{0,1\}^C$ to every step $k$ as a supervision.
Loss from a single step is computed by a cross entropy between the answer probability ${\bf a}^k$ and the ground-truth answer ${\bf y}$.
Loss from each step is simply aggregated to compute the overall loss.
Given image $I_n$, question $q_n$, and ground-truth label ${\bf y}_n$ of the $n^{\rm th}$ training example, the joint loss function is formally given by
\begin{equation}
\mathcal{L} =
       \frac{1}{N}
        \sum_{n=1}^{N}{
                 \sum_{k=1}^{K}{
                         -{\bf y}^\mathrm{T}_n \
                         \mathrm{log}
                         \left(
                             {\bf a}^k_n
                         \right)
                 }
         },
\label{eq:overall_loss}
\end{equation}
where 
\begin{equation}
{\bf a}^k_n =
                             \mathrm{Answer}_k  \left(  {\bf f}_I  (  I_n  ; {\bm \theta}_I )  ,  
                                                                     {\bf f}_q ( q_n  ; {\bm \theta}_q ) ,  
                                                                     {\bf h}^{k-1}_\mathrm{mem} ;  
                                                                     {\bm \theta}_\mathrm{ans}
                                                            \right) .
\end{equation}
Note that ${\bm \theta}_I$ and ${\bm \theta}_q$ denote parameters for image and question encoder, respectively.
The model parameter for the answering module is given by ${\bm \theta}_\mathrm{ans} = \{ {\bm \theta}_\mathrm{sub} , {\bm \theta}_\mathrm{att}, 
{\bm \theta}_\mathrm{soft}, {\bm \theta}_\mathrm{LSTM} \}$.

We train the model end-to-end by backpropagation, where the objective function based on cross entropy in Eq.~\eqref{eq:overall_loss} is minimzed.
Given an image $I_n$ and a question $q_n$, image encoder and question encoder extract features, ${\bf f}_I$ and ${\bf f}_q$, respectively.
The extracted features are given to the individual unfolded answering units.
As the answering module itself is a recurrent neural network, we compute the gradients of all the parameters in ${\bm \theta}_\mathrm{ans}$ by backpropagation through time (BPTT).
Note that the backpropagated loss to the inputs of answering module, ${\bf f}_I$ and ${\bf f}_q$, is used to learn the parameters of the image and question encoders, ${\bm \theta}_I$ and ${\bm \theta}_q$. 

When the network is trained with the joint loss from multiple steps as in Eq.~\eqref{eq:overall_loss}, we observe that models with multiple steps generally overfit to training data easily and show lower validation/testing accuracy than the model in the first step as illustrated in Figure~\ref{fig:multistep_overfit}. 
Note that this observation also coincides with~\cite{SAN}, which states that using three or more attention does not improve the performance any further.
We believe that the answering units in later steps have more model capacity and can fit to training data better while it loses the generality of models.
Such overfitting tendency may ruin the models in the earlier steps since the losses in the later steps are propagated backwards until the first answering unit.
To circumvent this issue, we stop backpropagating losses from the overfitted answering units progressively as soon as we identify their overfitting.
In practice, we validate the model in every epoch and early-stop training from an overfitted answering unit if the validation accuracy of the answering unit drops more than a predefined threshold from its maximum value.
According to our observation, the answering units in the later steps typically are terminated first and the first answering unit always manages to stay alive.
If validation dataset is not available, we early-stop training based on the formula for convenience, which is empirically determined as
\begin{equation}
t^k_\mathrm{stop}  =  
      \mathrm{round} \hspace{-0.05cm} \left[
                                      \left( t_\mathrm{max} - t_\mathrm{min} \right) 
                                      \frac{   e^{\lambda (K-k)} - 1    }
                                            {   e^{\lambda (K-1)} - 1   }
                                      +
                                      t_\mathrm{min}
                              \right], \hspace{-0.05cm}
\label{eq:early-stop}                             
\end{equation}
where $t_\mathrm{min}$ is the number of epochs before the first early stopping,  $t_\mathrm{max}$ is the total number of epochs for training and $t^k_\mathrm{stop}$ is the number of epochs before training the $k^{\rm th}$ answering unit is terminated. 
The early stopping is scheduled by controlling the configuration parameter $\lambda$.

\subsection {Testing}
In testing time, we predict answers only from the first answering unit in our model since it has better generalization accuracy in our experiments.
Although we believe that some complex tasks can be solved better in the later steps, it is very difficult to know which step is optimal to find the correct solution.
Hence, selecting one of the answers from multiple answering units is not feasible in practice.
Also, the ensemble of the answers from multiple units is not particularly better than the solution from the first unit. 
Note that, since the losses from the later steps are always backpropagated until the first answering module, it gradually learns how to solve more complex problems and does not tend to overfit by employing early-stopping strategy.

\subsection {Comparison to Memory Network based Approaches}
At first glance, the proposed method resembles the VQA methods such as~\cite{SAN,DMNQA}, which are based on memory network~\cite{Memnet}.
However, we introduce a novel training strategy with early-stopping and a simple testing method with a single-step inference.
This decision is based on our observations in VQA problems, but it is contradictory to \cite{SAN,DMNQA}.
The previous works~\cite{SAN,DMNQA} state that multi-step training and testing without parameter sharing is advantageous, but our results suggest that multi-step training with parameter sharing and a single-step testing may be better in practice.
This is partly supported by Figure~\ref{fig:multistep_overfit} although the experiment setting is not exactly identical.
Also, our experiment supports the claim that benefit from multi-step training with weight sharing is larger than that from multi-step training and testing without sharing weights.

\section{Experiment}
\label{sec:experiment}

\subsection{Dataset and Evaluation Metric}
We train and test the proposed network in VQA dataset~\cite{VQA}, which borrows images from MSCOCO dataset~\cite{Mscoco} and collects questions and answers via Amazon Mechanical Turk.
The dataset consists of 248,349 questions for training, 121,512 questions for validation, and 244,302 questions for testing.
For each image, 3 questions are asked and 10 independent answers are given to each question.
There are two test datasets; we typically use {\it test-dev} split for the control experiments and {\it test-standard} for comparison with external algorithms.

Two tasks are defined on VQA dataset: open-ended task and multple-choice task.
The model has to predict an answer for an open-ended question without knowing predefined candidate answers, but select one of 18 candidate answers in multiple-choice task
In both cases, the answers given by a model are evaluated by the following metric reflecting human consensus:
\begin{equation}
\mathrm{Acc}_\mathrm{VQA} (ans) = \mathrm{min} \left(   \frac{     \text{\#humans that said }  ans    }{3}   , 1         \right)
\end{equation}
where the score for a question is proportional to the number of matches with ground-truth answers, and the tested model receives full credit for each question if at least $3$ people agree to the predicted answer. 

\subsection{Implementation Details}

We use VGG-16 net~\cite{Vgg16} and ResNet-101~\cite{Resnet} as image encoders.
After rescaling input images to $448 \times 448$, we extract the feature maps from the last pooling layer in VGG or the layer below global average pooling layer in ResNet.
The question features are given by a $2$-layer LSTM\footnote{\url{https://github.com/VT-vision-lab/VQA\_LSTM\_CNN}}; the final hidden and cell states in both layers are concatenated to be used as a question feature.
The dimensionality of the attended image feature and the subtask feature are both 512.
For better generalization, we apply dropout with rate of $0.5$ to the features for input images and questions in each answering unit independently.
The vocabulary size of our model is $14,772$ and the rest of words are converted to UNK tokens.
The set of possible answers ($\Omega$ in Eq.~\eqref{eq:objective}) contains $1,000$ answers with the highest frequencies among all answers in the training dataset.

We set the number of steps $K$ to $8$ for multi-step training, and determine the parameters in Eq.~\eqref{eq:early-stop}, $t_\mathrm{max}$, $t_\mathrm{min}$, and $\lambda$, using validation set.
We use Adam~\cite{Adam} for optimization.
Learning rate for question encoder and answering modules are set to $3 \times 10^{-3}$ and $3 \times 10^{-4}$, respectively, and the image encoder is not fine-tuned.
Both learning rates are decayed in every epoch by the factor of $0.9$.
Additionally, we inject random noises sampled from Gaussian distribution to the gradient as suggested in~\cite{Noisygradient}, where $\eta$ is the number of training iterations.
To alleviate the exploding gradient problem, we limit the magnitude of gradient vector to 0.1 by normalization.
To facilitate reproduction, we make our code publicly available\footnote{\url{http://cvlab.postech.ac.kr/research/rau\_vqa/}}.
\begin{table}[!t]\scriptsize
\centering
\caption{
Single model performance on the VQA test-dev dataset of all compared algorithms including the variations of our algorithm.
Asterisk (*) denotes the concurrent submission with this paper.
}
    \begin{tabular}
        {
           @{}L{3.45cm}@{}   |   
           @{}C{0.6cm}@{}@{}C{0.6cm}@{}@{}C{0.6cm}@{}@{}C{0.7cm}@{}     |
           @{}C{0.6cm}@{}@{}C{0.6cm}@{}@{}C{0.6cm}@{}@{}C{0.7cm}@{}
        }
 \multirow{2}{*}{} &\multicolumn{4}{c|}{Open-Ended}      & \multicolumn{4}{c}{Multiple-Choice} \\
                                              & All     &Y/N    & Num  & Others        & All     &Y/N    & Num & Others \\

         \hline
         BoW Q~\cite{VQA}         &48.1   &75.7   &36.7   &27.1            &53.7  &75.7   &37.1  &38.6  \\
         I~\cite{VQA}                   &28.1   &64.0   &00.4   &03.8            &30.5  &69.9   &00.5  &03.8  \\
         Bow Q+I~\cite{VQA}       &52.6   &75.6   &33.7   &37.4            &59.0  &75.6   &34.4  &50.3  \\
         LSTM Q~\cite{VQA}       &48.8   &78.2   &35.7   &26.6            &54.8  &78.2   &36.8  &38.8  \\
         LSTM Q+I~\cite{VQA}    & 53.7  &78.9   &35.2   &36.4            &57.2  &79.0   &35.8  &43.4  \\
         \hline
         \cite{VQA}&57.8&80.5&36.8&43.1     &62.7  &80.5   &38.2  &53.0 \\
         \cite{NMN}&54.8   &77.7   &37.2   &39.3            & -        & -       & -       &  -     \\
         \cite{zhou2015simple}&55.7&76.6&35.0&42.6       &61.7   &76.7   &37.1  &54.4  \\
         \cite{Wheretolook}&- &- &- &-                   &62.4   &77.6   &34.3  &55.8 \\
         \cite{DPPnet}   &57.2   &80.7   &37.2   &41.7            &62.5   &80.8   &38.9  &52.2  \\
         \cite{NMN2}     &57.9   &80.5   &37.4&43.1                & -       & -        & -       & -      \\
         \cite{SAN}            &58.7   &79.3   &36.6   &46.1            & -       & -        & -       & -      \\
         \cite{FreeformVQAKB}&59.2&81.0&38.4&45.2               & -       & -        & -       & -      \\         
         \cite{DMNQA}     &60.3   &80.5   &36.8    &48.3            & -       & -        & -       & -      \\
         \hline
         \cite{kim2016multimodal} ResNet*
         &61.7&82.3&38.8&49.3& -       & -        & -       & -    \\ 
         \cite{lu2016hierarchical} ResNet*
         &61.8&79.7&38.7&51.7
         &65.8&79.7&40.0&59.8\\
         \cite{fukui2016multimodal} ResNet*
         &64.7&82.5&37.6&55.6
         &69.1&-&-&-\\
         \hline
         Ours\_SS                     &59.0   &78.3   &37.1    &47.6            &64.0  &78.4   &38.2   &57.5 \\
         Ours\_MS                    &61.0   &81.3   &36.8    &49.1            &65.8  &81.3   &38.9   &58.7 \\
         Ours\_FULL                 &61.3   &81.5   &37.0   &49.6             &66.1  &81.5   &39.5   &58.9 \\
         \hline
         Ours\_ResNet              &63.3&81.9&39.0&53.0&67.7&81.9&41.1&61.5\\
         \hline
    \end{tabular}
\label{tab:testdev} 
\end{table}

\begin{figure}[t]
    \centering
    \includegraphics[width=0.7\linewidth] {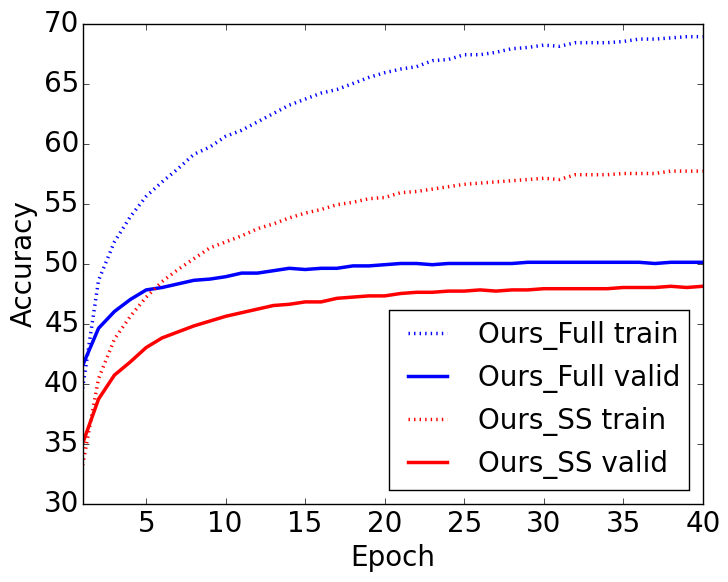}
    \vspace{-0.3cm}
    \caption{
        Training and validation accuracy of {\sf \small Ours\_SS} and {\sf \small Ours\_Full}.
    }
    \vspace{-0.3cm}
    \label{fig:trainval_curve}
\end{figure}

\begin{figure}[!t]
\centering
\includegraphics[height=0.059\linewidth] {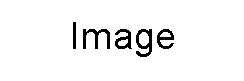}
\includegraphics[height=0.059\linewidth] {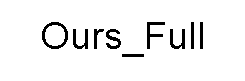}
\includegraphics[height=0.059\linewidth] {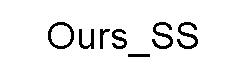} 
\includegraphics[height=0.059\linewidth] {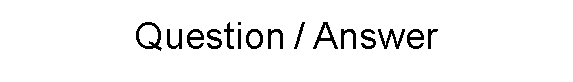} \\
\vspace{-0.12cm}
\includegraphics[height=0.18\linewidth] {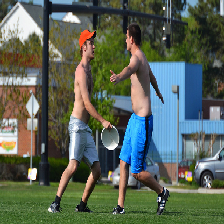}
\includegraphics[height=0.18\linewidth] {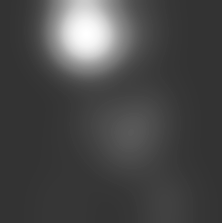}
\includegraphics[height=0.18\linewidth] {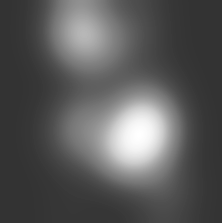} 
\includegraphics[height=0.18\linewidth] {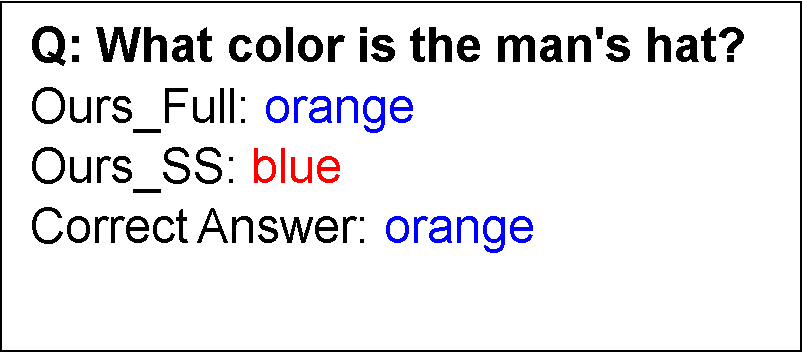} \\
\includegraphics[height=0.18\linewidth] {./qr_9_4235192.png}
\includegraphics[height=0.18\linewidth] {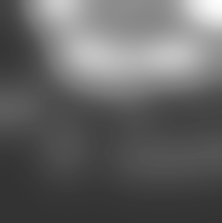}
\includegraphics[height=0.18\linewidth] {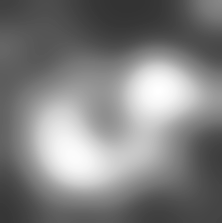}
\includegraphics[height=0.18\linewidth] {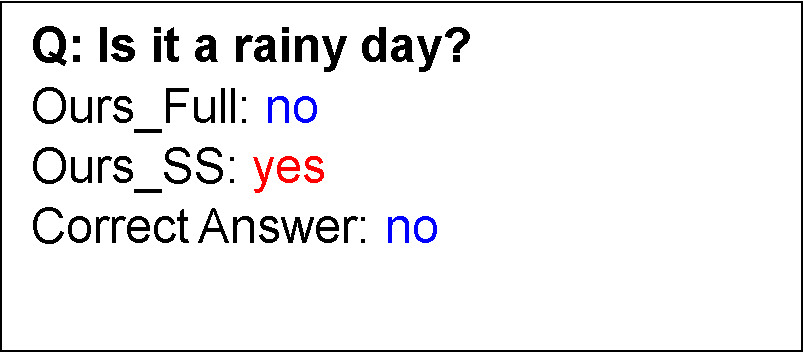} \\
\includegraphics[height=0.18\linewidth] {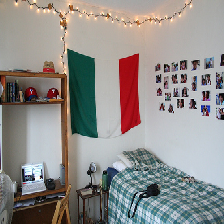}
\includegraphics[height=0.18\linewidth] {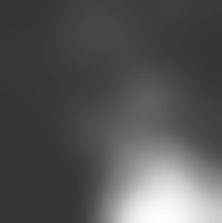}
\includegraphics[height=0.18\linewidth] {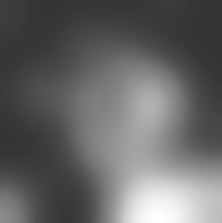}
\includegraphics[height=0.18\linewidth] {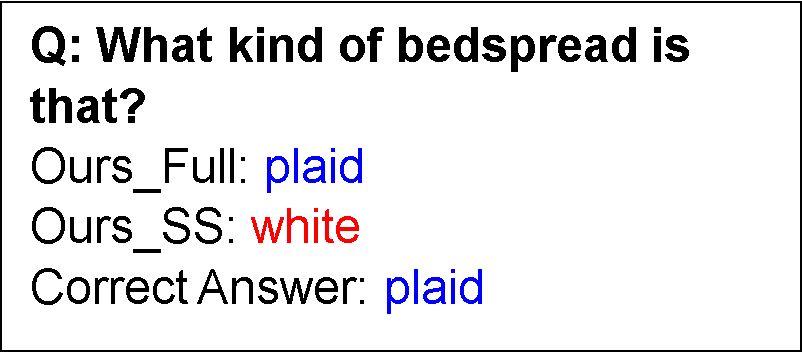} \\
\includegraphics[height=0.18\linewidth] {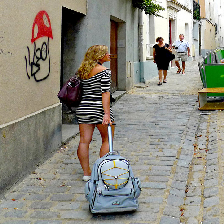}
\includegraphics[height=0.18\linewidth] {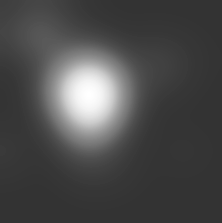}
\includegraphics[height=0.18\linewidth] {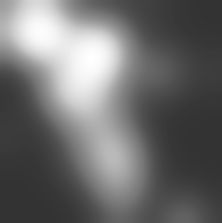}
\includegraphics[height=0.18\linewidth] {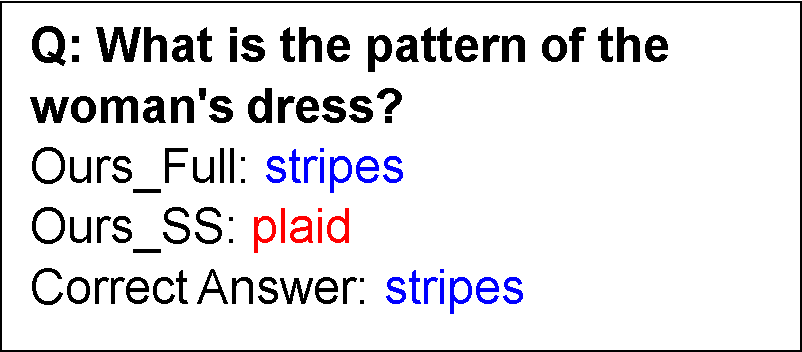} \\
\includegraphics[height=0.18\linewidth] {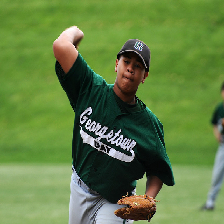}
\includegraphics[height=0.18\linewidth] {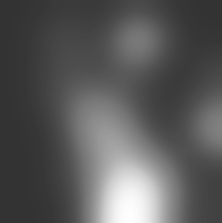}
\includegraphics[height=0.18\linewidth] {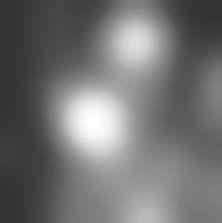}
\includegraphics[height=0.18\linewidth] {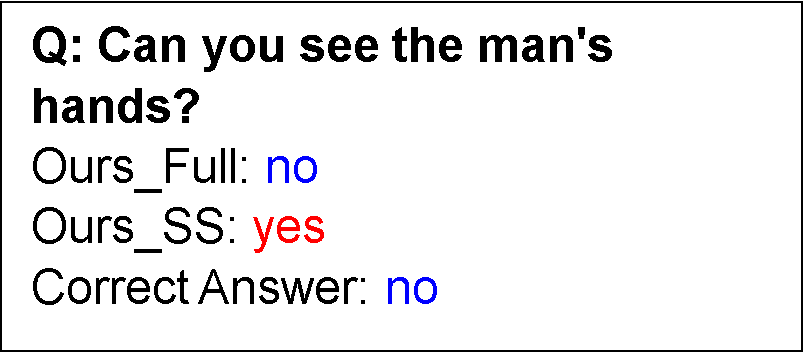} \\
\vspace{-0.2cm}
\caption{
Qualitative comparisons of attention between {\sf \small Ours\_Full} and {\sf \small Ours\_SS}.
}
\vspace{-0.3cm}
\label{fig:qualitative}
\end{figure}

\begin{table}[t]\scriptsize
\centering
\caption{
Comparisons of single model performance in the VQA test-standard.
Asterisk (*) denotes the concurrent submission with this paper.
}
    \begin{tabular}[b]
        {
           @{}L{3.45cm}@{}   |   
           @{}C{0.6cm}@{}@{}C{0.6cm}@{}@{}C{0.6cm}@{}@{}C{0.7cm}@{}     |
           @{}C{0.6cm}@{}@{}C{0.6cm}@{}@{}C{0.6cm}@{}@{}C{0.7cm}@{}           
        }
 \multirow{2}{*}{} &\multicolumn{4}{c|}{Open-Ended}      & \multicolumn{4}{c}{Multiple-Choice} \\
                                              & All     &Y/N    & Num  & Others        & All     &Y/N    & Num & Others        \\
         \hline
         Human~\cite{VQA}        &83.3   &95.8   &83.4    &72.7             &-        &-         &-        &-                 \\ 
         \hline
         \cite{VQA}&58.2&80.6&36.5&43.7     &63.1    &80.6    &37.7  &53.6          \\
         \cite{NMN}&55.1  & -        & -         & -                 & -        & -        & -       & -              \\
         \cite{zhou2015simple}&55.9&76.8&35.0&42.6        &62.0   &76.9    &37.3   &54.6          \\
         \cite{Wheretolook}&-   & -   & -   & -          &62.4    &77.2   &33.5   &56.1          \\
         \cite{DPPnet}  &57.4   &80.3    &36.9    &42.2            &62.7   &80.4    &38.8  &52.8          \\
         \cite{NMN2}    &58.0   & -        & -         & -                & -       & -         & -       & -               \\
         \cite{SAN}           &58.9   & -        & -         & -                & -       & -         & -       & -               \\
         \cite{FreeformVQAKB}&59.4&81.1&37.1   &45.8            & -       & -        & -       & -                \\          
         \cite{DMNQA}    &60.4   &80.4  &36.8  &48.3          & -       & -         & -       & -               \\
         \hline         
         \cite{kim2016multimodal} ResNet*
         &61.8&82.4&38.2&49.4
         &66.3&82.4&39.6&58.4\\ 
         \cite{lu2016hierarchical} ResNet*
         &62.1&-&-&-
         &66.1&-&-&-\\
         \hline
         Ours\_ResNet             &63.2&81.7&38.2&52.8&67.3&81.7&40.0&61.0        \\
         \hline
    \end{tabular}
\vspace{-0.7cm}
\label{tab:teststd} 
\end{table}

\subsection{Results and Analysis}
To ensure the effectiveness of the multi-step training and early-stopping strategy, we perform several control experiments.
The baseline is the proposed model without both components, which is referred to as {\sf \small Ours\_SS}.
We also test the model trained with multi-step training but without early-stopping strategy, {\sf \small Ours\_MS}, and our full model denoted by {\sf \small Ours\_FULL} employ both strategies for training.
We use the image features extracted from VGG-16 net~\cite{Vgg16} for all the control experiments.

Table~\ref{tab:testdev} illustrates the single model performance of various approaches trained without data augmentation.
The proposed algorithm, {\sf \small Ours\_FULL} outperforms the models based on multi-step attention~\cite{DMN,SAN,lu2016hierarchical} in the VQA dataset using the same image encoder.
The best result is achieved by the work based on the multimodal compact bilinear pooling~\cite{fukui2016multimodal}.
We believe that the multimodal compact bilinear pooling can be integrated within the proposed model, but we leave it as a future work.

Note that the performance gain from {\sf \small Ours\_SS} to {\sf \small Ours\_FULL} is about $2.3$, which are significant in VQA context.
Both training schemes turn out to be useful for performance improvement.
These results are interesting because multi-step training is effective even for a single-step prediction and early-stopping strategy improves accuracy as long as all active steps are free from overfitting.
Since we schedule early-stopping based on the formula from simple empirical study, the accuracy may be further improved with more sophisticated validation.

Figure~\ref{fig:trainval_curve} illustrates training and validation accuracy of {\sf \small Ours\_FULL} and {\sf \small Ours\_SS}.
The proposed training strategy denoted by {\sf \small Ours\_FULL} outperforms {\sf \small Ours\_SS} consistently even though they have the exactly same architecture for a single step prediction during testing.
%
Figure~\ref{fig:qualitative} presents predicted answers with attended regions.
Visualization of attention shows that {\sf \small Ours\_FULL} tends to focus on critical regions while {\sf \small Ours\_SS} are often distracted by irrelevant objects.
We believe that our training strategy helps the model learn to answer questions in appropriate steps in training while it enables the answering unit to implicitly solve a series of subtasks given a question in testing.


\subsection{Comparison with Stacked Attention Network}
The architecture of our single answering unit is similar to the stacked attention network (SAN)~\cite{SAN}.
The difference lies in how to handle questions that require multiple steps to answer.
As the number of required steps varies across questions and it is difficult to figure out the desirable number, SAN uses the same fixed number of steps for training and testing, which might result in overfitting observed in Figure~\ref{fig:multistep_overfit}.
Instead, we provide multiple supervision in our network, but hope a single answering unit (the first one, in practice) to learn the best model for evaluation in testing.
The comparison between the SAN and the proposed method in Table~\ref{tab:testdev} clearly shows the effectiveness of our method.

\subsection{Experimental Results with ResNet}
We can improve the accuracy of a VQA system by using a better image encoder.
Hence, we train the model with image features extracted from ResNet-101~\cite{Resnet}, and this model is denoted by {\sf \small Ours\_ResNet}.
The model is trained with same hyper-parameters with {\sf \small Ours\_FULL} in Table~\ref{tab:testdev}, and evaluated both in VQA test-dev and test-standard.
The results from {\sf \small Ours\_ResNet} is presented in Table~\ref{tab:testdev} and \ref{tab:teststd}, where we observe that ResNet improves performance substantially.

\section{Conclusion}
\label{sec:conclusion}
We proposed a VQA algorithm based on a recurrent deep neural network, which is trained by minimizing the joint loss from all answering units.
The answering units share model parameters while the outputs from the units in the later steps depend on the results from the ones in the earlier steps.
To maximize performance, we introduce an early stopping training strategy, where individual answering units are disregarded in training as soon as they start to overfit to the training set.
Only the first answering unit is employed for inference since it is the model learned from all the multiple answering units without overfitting.
The proposed architecture illustrates the outstanding performance in the standard VQA dataset without data augmentation.
We believe that our algorithm has great potential to be used as a general framework for VQA problems by replacing our answering unit with any other networks.

{\small
\bibliographystyle{aaai}
\bibliography{recur_ans_unit_bib}
}


\clearpage
\thispagestyle{empty}
\onecolumn

\begin{center}

\textbf{{\\\vspace{1.2cm} \LARGE Training Recurrent Answering Units  with Joint Loss Minimization for VQA}} \\
\vspace{0.2cm}
\textbf{{\large Supplementary Document}} \\
\vspace{1.4cm}

\end{center}

\begin{multicols}{2}

This document provides our implementation details and additional results that could not be accommodated in the main paper due to space limitation.

\section{Implementation Details}
In this section, we provide implementation details for each components of answering units described in Eq.~(2) to (7) of the main paper.
Without loss of generality, we describe the architecture of the $k^{\rm th}$ answering unit.

\subsection{Subtask Module}
Subtask module in Eq.~(3) of the main paper generates subtask feature ${\bf f}_\mathrm{sub} \in \mathbb{R}^{S}$ from a question feature ${\bf f}_q \in \mathbb{R}^{Q}$ and the previous memory state ${\bf h}^{k-1}_\mathrm{mem} \in \mathbb{R}^{M}$. 
This operation is given by
\begin{equation}
    {\bf f}_\mathrm{sub}^k
        = \tanh ( {\bf W}_q   {\bf f}_q   +   {\bf W}_h    {\bf h}^{k-1}_\mathrm{mem} + {\bf b}_\mathrm{sub} ),
\end{equation}
where ${\bf W}_q \in \mathbb{R}^{S \times Q}$, ${\bf W}_h \in \mathbb{R}^{S \times M}$ and ${\bf b}_\mathrm{sub} \in \mathbb{R}^{S}$ are weight parameters.

\subsection{Attention Module}
Attention module in Eq.~(4) of the main paper generates the attention probability map ${\bf f}_\mathrm{loc}^k \in \mathbb{R}^{L}$  and attended feature ${\bf f}_\mathrm{att}^k \in \mathbb{R}^{S}$ based on a soft attention mechanism.
The computation of ${\bf f}_\mathrm{loc}^k$ and ${\bf f}_\mathrm{att}^k$ requires embedding of input image feature map ${\bf f}_I \in \mathbb{R}^{P \times L}$ (with $P$ channels and $L$ locations), and the embedded feature map ${\bf g}_I$ is given by
\begin{equation}
    {\bf g}_I =  \tanh ( {\bf W}_I {\bf f}_I + {\bf b}_I ),
\end{equation}
where ${\bf W}_I \in \mathbb{R}^{S \times P}$ and ${\bf b}_I \in \mathbb{R}^{S}$ are weight parameters.

The procedure to compute attention probability map ${\bf f}_\mathrm{loc}^k \in \mathbb{R}^{L}$ is composed of the following three steps.
First, we compute pre-attention score ${\bf \alpha}_k \in \mathbb{R}^{L}$ based on the embeded image feature map ${\bf g}_I \in \mathbb{R}^{S \times L}$ and the subtask feature ${\bf f}_\mathrm{sub}^k$ by the following equation:
\begin{equation}
    {\bf \alpha}_k^T =        
        {\bf W}_\alpha^1 \tanh \left(
            {\bf W}_\alpha^2 {\bf g}_I +
            {\bf W}_\alpha^3 {\bf f}_\mathrm{sub}^k {\bf 1}^{T} +
            {\bf b}_\alpha {\bf 1}^{T}
        \right) ,
\end{equation}
where ${\bf W}_\alpha^1 \in \mathbb{R}^{L \times A}$, ${\bf W}_\alpha^2 \in \mathbb{R}^{A \times S}$, ${\bf W}_\alpha^3 \in \mathbb{R}^{A \times S}$ and ${\bf b}_\alpha \in \mathbb{R}^{A}$ are weight parameters, and ${\bf 1} \in \mathbb{R}^{L}$ is one vector.
Next, we compute attention score $\beta_k \in \mathbb{R}^{L}$ by adding another pre-attention score extracted from the previous memory ${\bf h}^{k-1}_\mathrm{mem}$, which is given by
\begin{equation}
    \beta_k = \alpha_k + {\bf W}_\beta {\bf h}^{k-1}_\mathrm{mem} + {\bf b}_\beta,
\end{equation}
where ${\bf W}_\beta \in \mathbb{R}^{L \times M}$ and ${\bf b}_\beta \in \mathbb{R}^{L}$ are weight parameters.
Last, attention probability map ${\bf f}_\mathrm{loc}^k \in \mathbb{R}^{L}$ is given by applying softmax function to the attention score $\beta_k \in \mathbb{R}^{L}$ as described below:
\begin{equation}
    {\bf f}_\mathrm{loc}^k = \frac{  \mathrm{exp}  \left( \beta_k \right)  }{ {\bf 1}^T \mathrm{exp}  \left( \beta_k \right) }.
\end{equation}
Once ${\bf f}_\mathrm{loc}^k$ is obtained, attended feature ${\bf f}_\mathrm{att}^k \in \mathbb{R}^{S}$ is computed using embedded image feature ${\bf g}_I$ and attention probability map ${\bf f}_\mathrm{loc}^k$ as follws:
\begin{equation}
    {\bf f}_\mathrm{att}^k = {\bf g}_I {\bf f}_\mathrm{loc}^k .
\end{equation}

\subsection{Prediction Module}
Prediction module in Eq.~(5) of the main paper has two sub-module: LSTM and softmax classification.
As specified in the main paper, we use publicly available LSTM implementation.
Single layer LSTM is constructed and the dimensionality of LSTM hidden state is equal to that of memory state ${\bf h}^{k-1}_\mathrm{mem} \in \mathbb{R}^{S}$.
Input to the LSTM, denoted by ${\bf f}_\mathrm{join}^k$, is computed by
\begin{equation}
    {\bf f}_\mathrm{join}^k = {\bf f}_\mathrm{sub}^k +
                                         {\bf f}_\mathrm{att}^k +
                                         {\bf W}_\mathrm{join}{\bf f}_\mathrm{loc}^k +
                                         {\bf b}_\mathrm{join},
\end{equation}
where ${\bf W}_\mathrm{join} \in \mathbb{R}^{S \times L}$ and ${\bf b}_\mathrm{join} \in \mathbb{R}^{S}$ are weight parameters. 
Answer probability ${\bf a}^k \in \mathbb{R}^{C}$ is obtained by applying a softmax function, which is given by
\begin{align}
 {\bf a}^k = \frac{ \mathrm{exp} \left( {\bf f}_\mathrm{join}^k  + {\bf W}_s {\bf h}_\mathrm{mem}^k + {\bf b}_s  \right) }{ {\bf 1}^T \mathrm{exp} \left( {\bf f}_\mathrm{join}^k  + {\bf W}_s {\bf h}_\mathrm{mem}^k + {\bf b}_s \right) },
\end{align}
where ${\bf W}_s \in \mathbb{R}^{C \times S}$ and ${\bf b}_s \in \mathbb{R}^{C}$ are weight parameters.

\subsection{Network Parameters}
Network parameters used for the experiments are as follows:
$S=512$, $Q=1024$, $M=512$, $P=512$, $L=196$, $A=256$, and $C=1000$.

\section{Additional Results}
Figure~\ref{fig:qualitative} presents qualitative comparison between {\sf \small Ours\_Full} and {\sf \small Ours\_SS} on VQA validation dataset.

\begin{figure*}[!t]
\centering
\includegraphics[height=0.0295\linewidth] {./qr_Image.png}
\includegraphics[height=0.0295\linewidth] {./qr_Ours_Full.png}
\includegraphics[height=0.0295\linewidth] {./qr_Ours_SS.png} 
\includegraphics[height=0.0295\linewidth] {./qr_QuestionAnswer.png} 
\hspace{-0.15cm}
\includegraphics[height=0.0295\linewidth] {./qr_Image.png}
\includegraphics[height=0.0295\linewidth] {./qr_Ours_Full.png}
\includegraphics[height=0.0295\linewidth] {./qr_Ours_SS.png} 
\includegraphics[height=0.0295\linewidth] {./qr_QuestionAnswer.png} \\
\vspace{-0.12cm}
\includegraphics[height=0.09\linewidth] {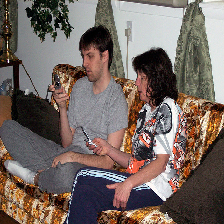}
\includegraphics[height=0.09\linewidth] {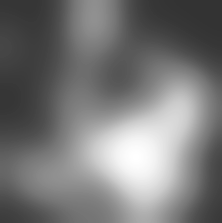}
\includegraphics[height=0.09\linewidth] {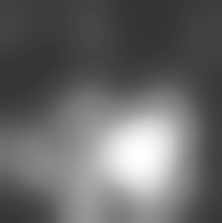}
\includegraphics[height=0.09\linewidth] {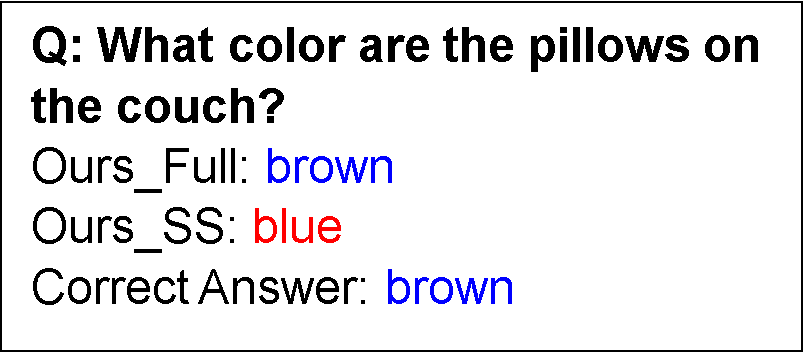} 
\includegraphics[height=0.09\linewidth] {./qr_3_608992.png}
\includegraphics[height=0.09\linewidth] {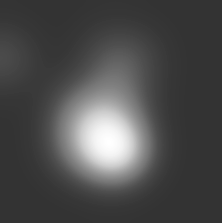}
\includegraphics[height=0.09\linewidth] {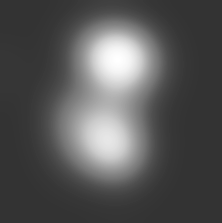} 
\includegraphics[height=0.09\linewidth] {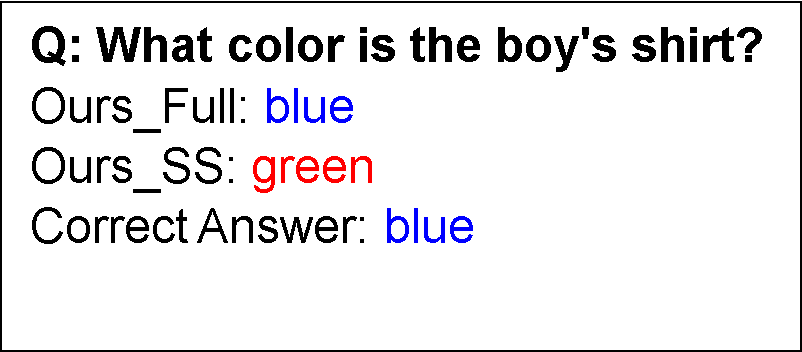} \\ 
\includegraphics[height=0.09\linewidth] {./qr_2_914692.png}
\includegraphics[height=0.09\linewidth] {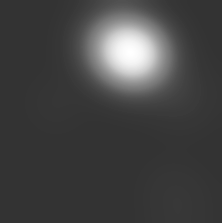}
\includegraphics[height=0.09\linewidth] {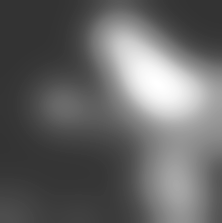} 
\includegraphics[height=0.09\linewidth] {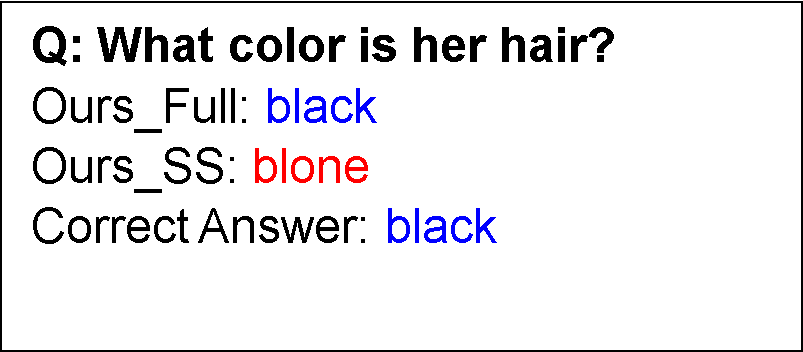} 
\includegraphics[height=0.09\linewidth] {./qr_8_3462321.png}
\includegraphics[height=0.09\linewidth] {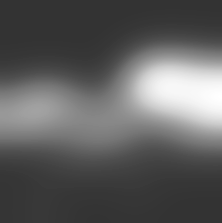}
\includegraphics[height=0.09\linewidth] {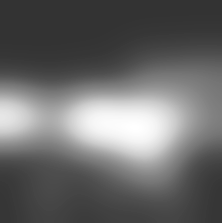} 
\includegraphics[height=0.09\linewidth] {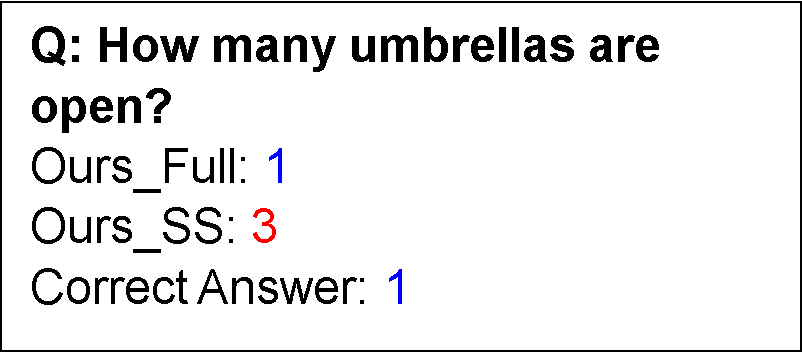} \\
\includegraphics[height=0.09\linewidth] {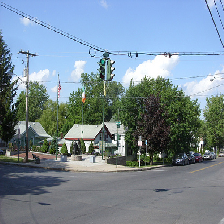}
\includegraphics[height=0.09\linewidth] {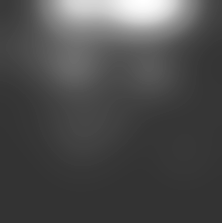}
\includegraphics[height=0.09\linewidth] {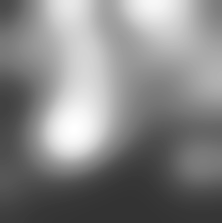}
\includegraphics[height=0.09\linewidth] {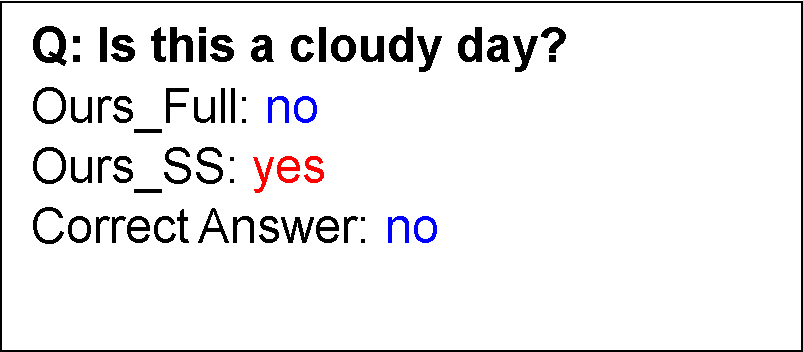} 
\includegraphics[height=0.09\linewidth] {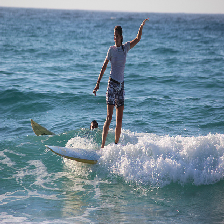}
\includegraphics[height=0.09\linewidth] {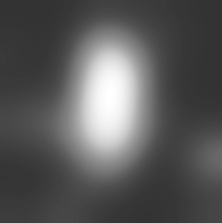}
\includegraphics[height=0.09\linewidth] {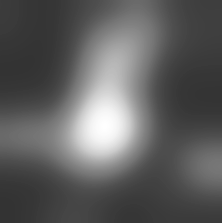}
\includegraphics[height=0.09\linewidth] {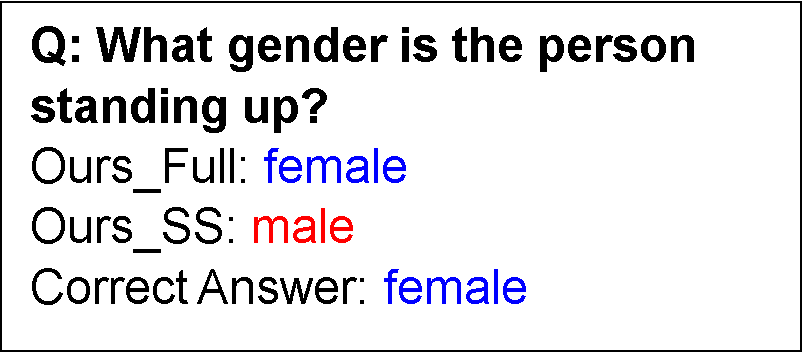} \\
\includegraphics[height=0.09\linewidth] {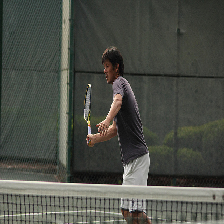}
\includegraphics[height=0.09\linewidth] {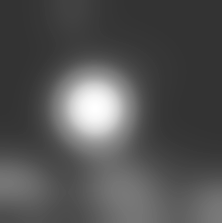}
\includegraphics[height=0.09\linewidth] {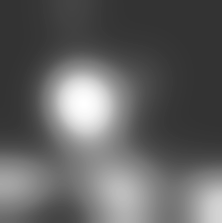}
\includegraphics[height=0.09\linewidth] {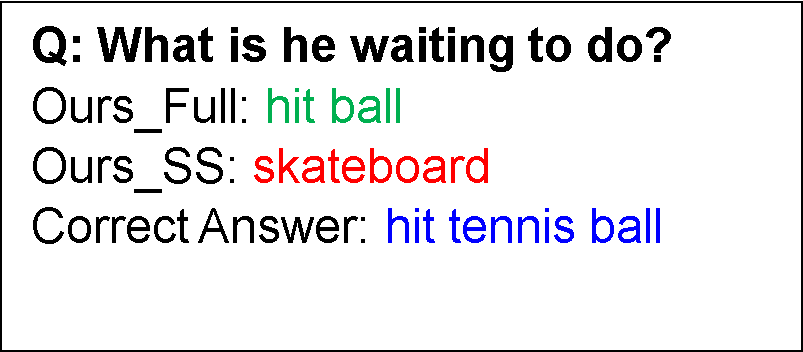} 
\includegraphics[height=0.09\linewidth] {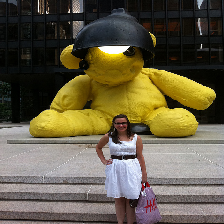}
\includegraphics[height=0.09\linewidth] {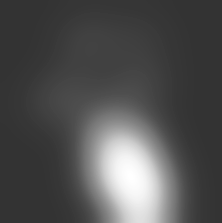}
\includegraphics[height=0.09\linewidth] {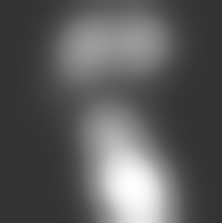} 
\includegraphics[height=0.09\linewidth] {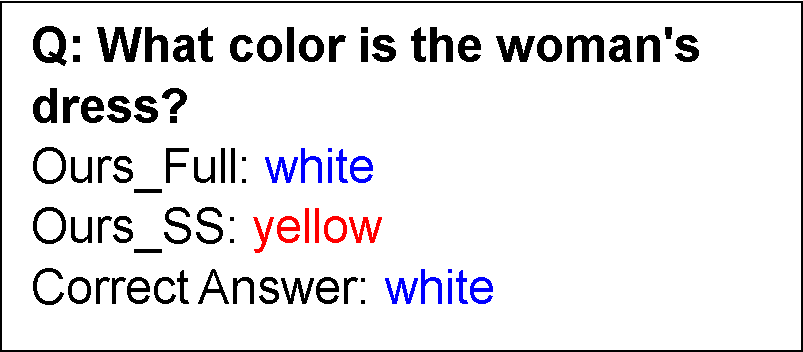} \\
\includegraphics[height=0.09\linewidth] {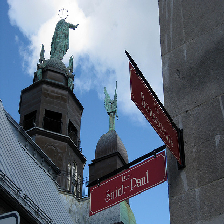}
\includegraphics[height=0.09\linewidth] {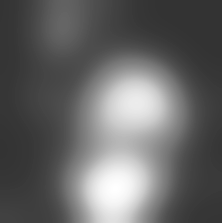}
\includegraphics[height=0.09\linewidth] {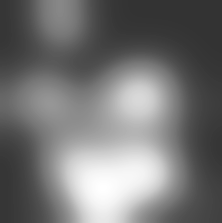}
\includegraphics[height=0.09\linewidth] {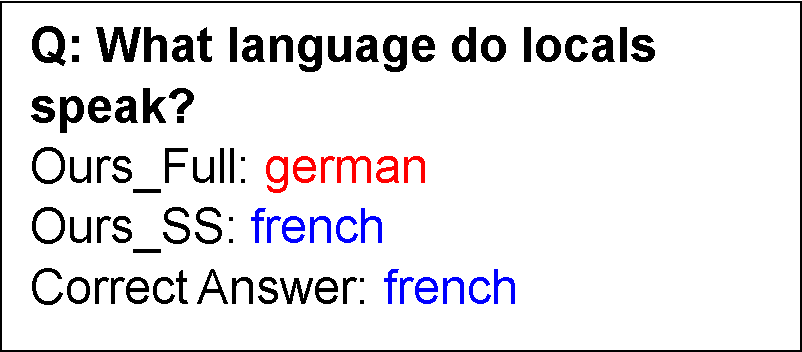} 
\includegraphics[height=0.09\linewidth] {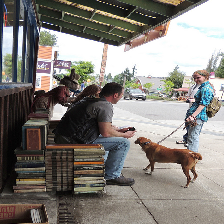}
\includegraphics[height=0.09\linewidth] {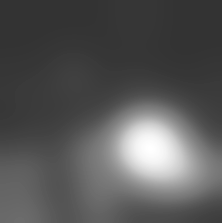}
\includegraphics[height=0.09\linewidth] {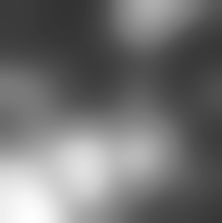}
\includegraphics[height=0.09\linewidth] {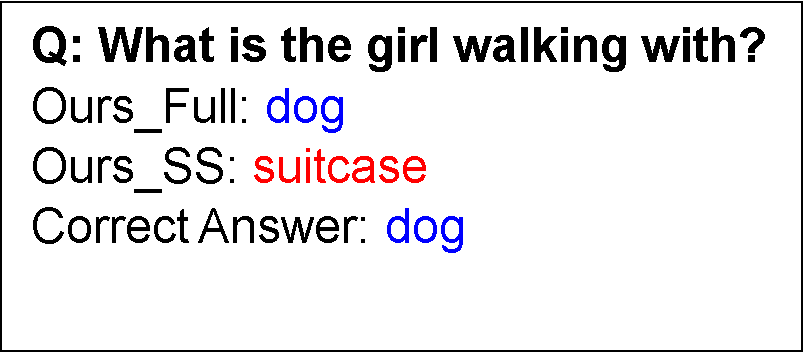}\\
\includegraphics[height=0.09\linewidth] {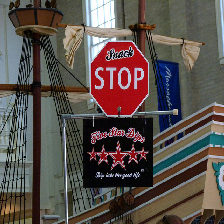}
\includegraphics[height=0.09\linewidth] {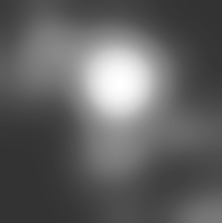}
\includegraphics[height=0.09\linewidth] {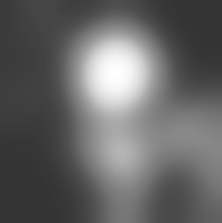} 
\includegraphics[height=0.09\linewidth] {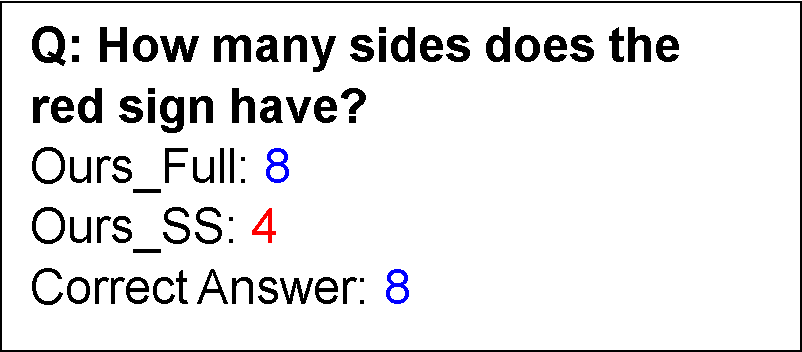} 
\includegraphics[height=0.09\linewidth] {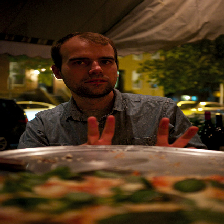}
\includegraphics[height=0.09\linewidth] {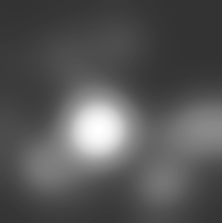}
\includegraphics[height=0.09\linewidth] {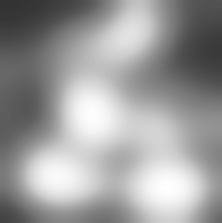} 
\includegraphics[height=0.09\linewidth] {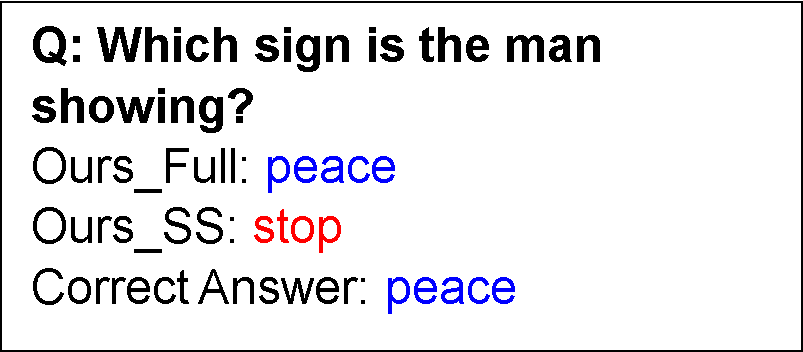} \\ 
\includegraphics[height=0.09\linewidth] {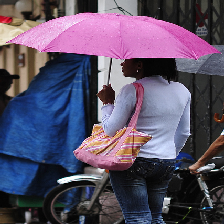}
\includegraphics[height=0.09\linewidth] {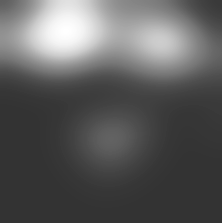}
\includegraphics[height=0.09\linewidth] {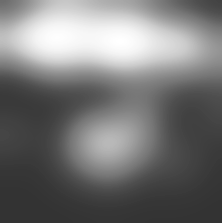} 
\includegraphics[height=0.09\linewidth] {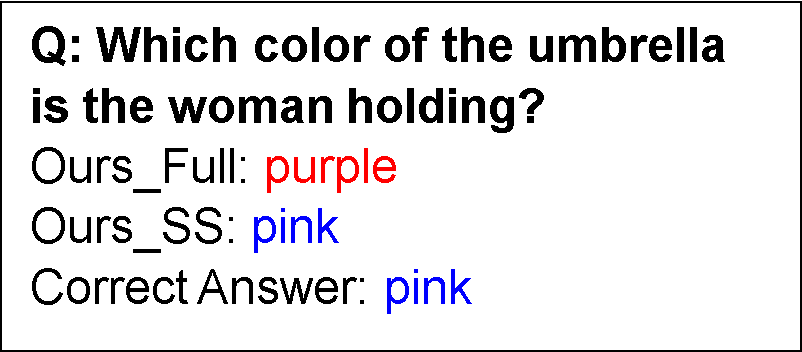} 
\includegraphics[height=0.09\linewidth] {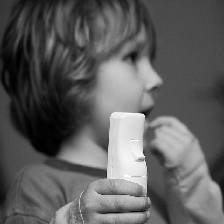}
\includegraphics[height=0.09\linewidth] {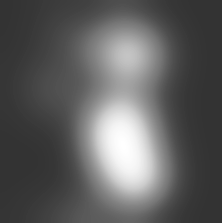}
\includegraphics[height=0.09\linewidth] {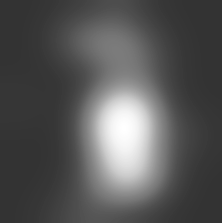} 
\includegraphics[height=0.09\linewidth] {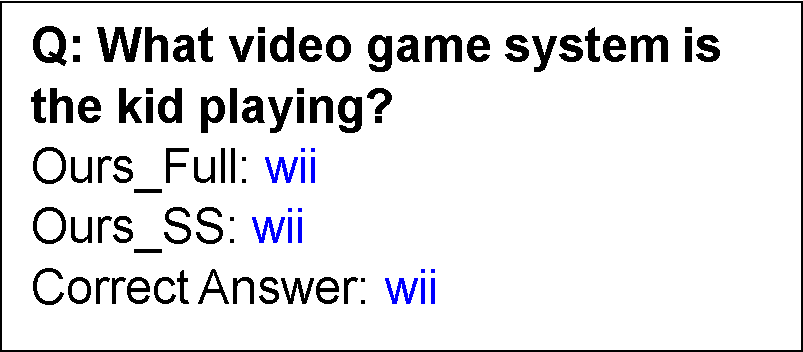} \\
\includegraphics[height=0.09\linewidth] {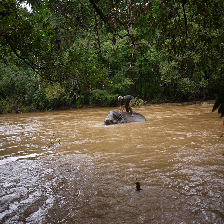}
\includegraphics[height=0.09\linewidth] {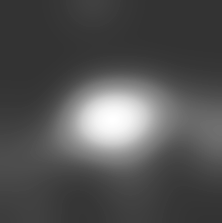}
\includegraphics[height=0.09\linewidth] {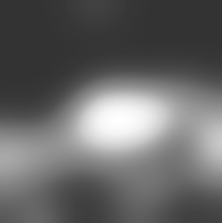}
\includegraphics[height=0.09\linewidth] {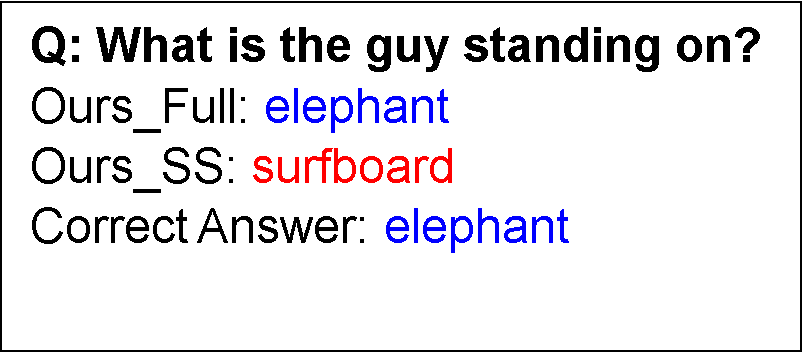} 
\includegraphics[height=0.09\linewidth] {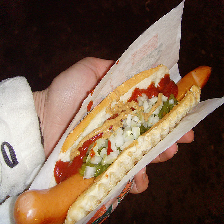}
\includegraphics[height=0.09\linewidth] {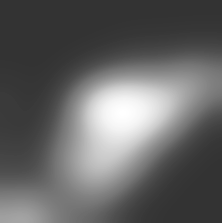}
\includegraphics[height=0.09\linewidth] {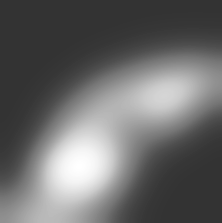}
\includegraphics[height=0.09\linewidth] {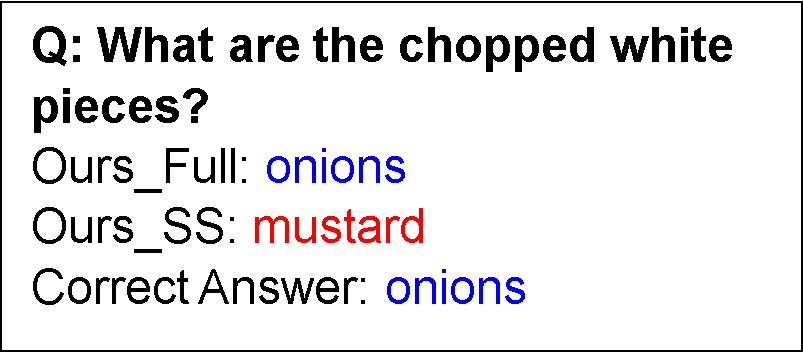} \\
\includegraphics[height=0.09\linewidth] {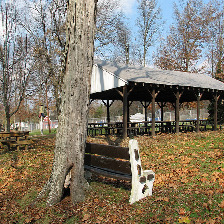}
\includegraphics[height=0.09\linewidth] {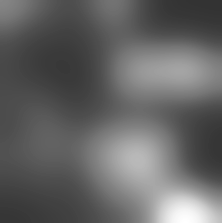}
\includegraphics[height=0.09\linewidth] {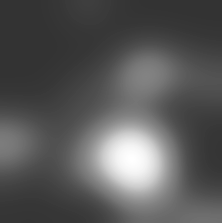}
\includegraphics[height=0.09\linewidth] {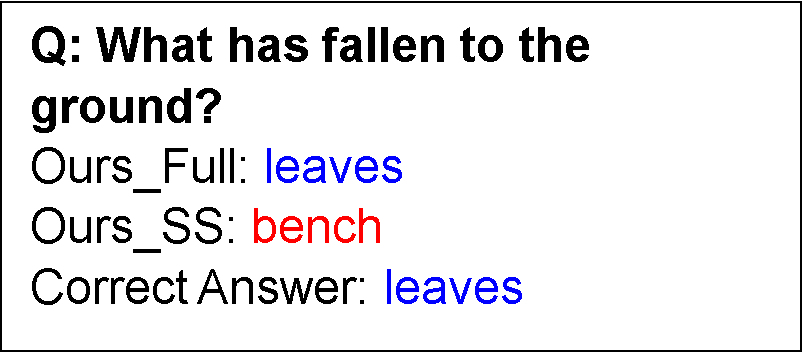} 
\includegraphics[height=0.09\linewidth] {./qr_27_5011422.png}
\includegraphics[height=0.09\linewidth] {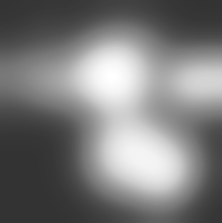}
\includegraphics[height=0.09\linewidth] {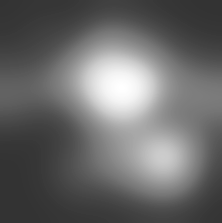}
\includegraphics[height=0.09\linewidth] {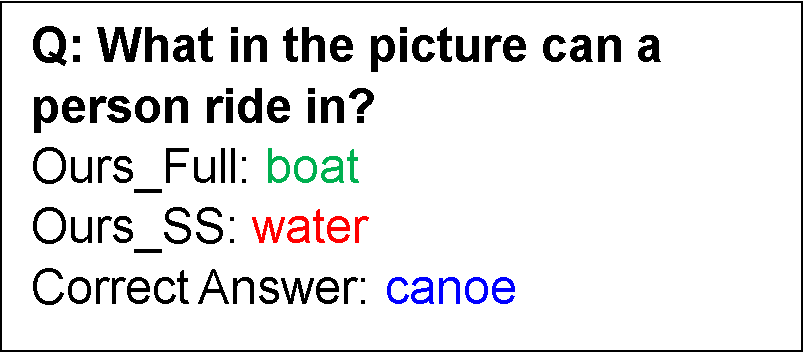} \\
\includegraphics[height=0.09\linewidth] {./qr_15_102192.png}
\includegraphics[height=0.09\linewidth] {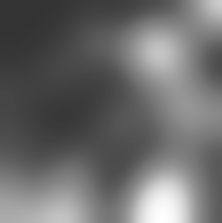}
\includegraphics[height=0.09\linewidth] {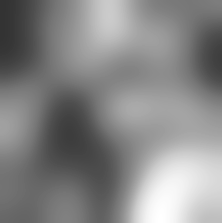}
\includegraphics[height=0.09\linewidth] {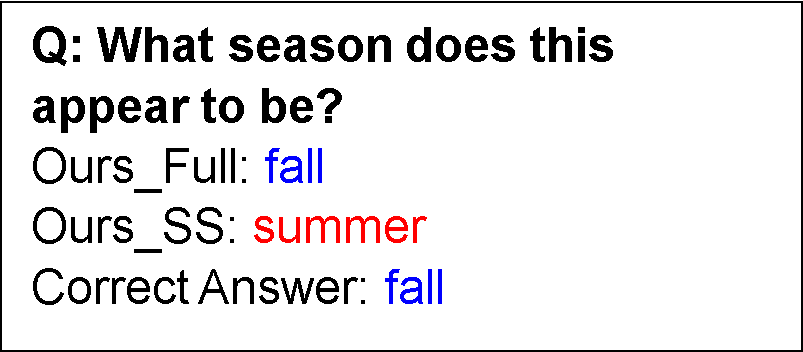} 
\includegraphics[height=0.09\linewidth] {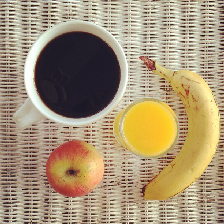}
\includegraphics[height=0.09\linewidth] {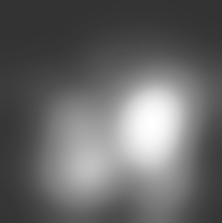}
\includegraphics[height=0.09\linewidth] {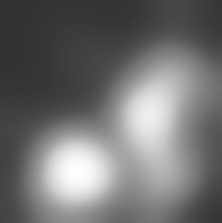}
\includegraphics[height=0.09\linewidth] {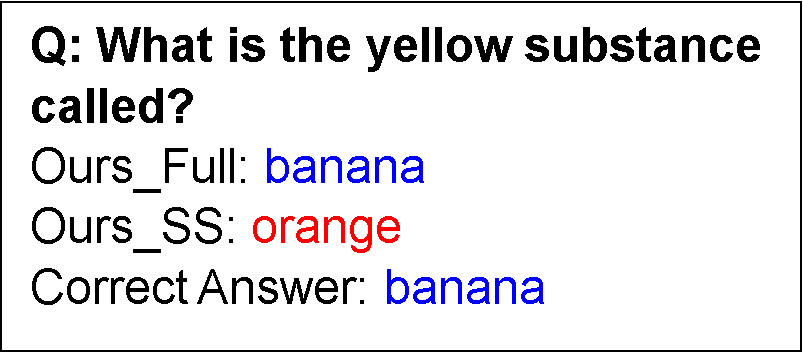} \\
\includegraphics[height=0.09\linewidth] {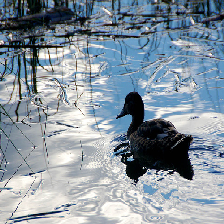}
\includegraphics[height=0.09\linewidth] {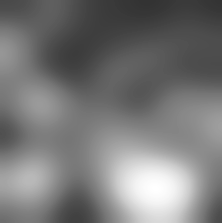}
\includegraphics[height=0.09\linewidth] {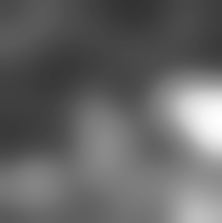}
\includegraphics[height=0.09\linewidth] {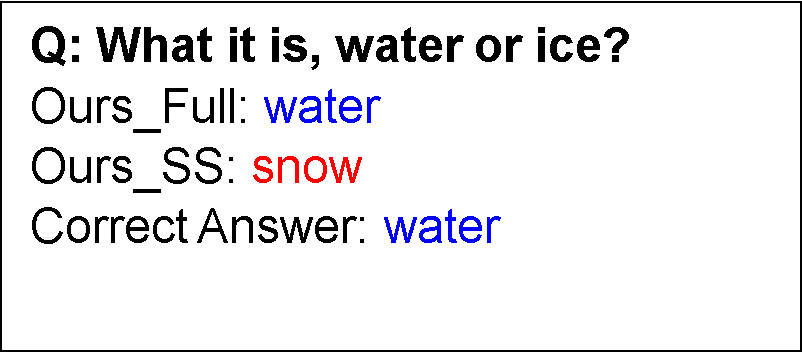} 
\includegraphics[height=0.09\linewidth] {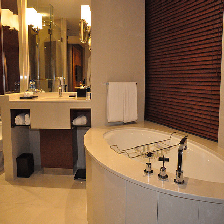}
\includegraphics[height=0.09\linewidth] {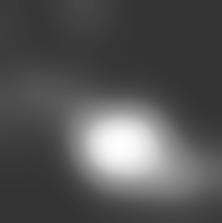}
\includegraphics[height=0.09\linewidth] {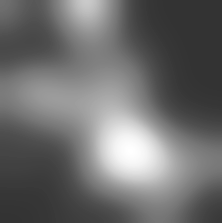}
\includegraphics[height=0.09\linewidth] {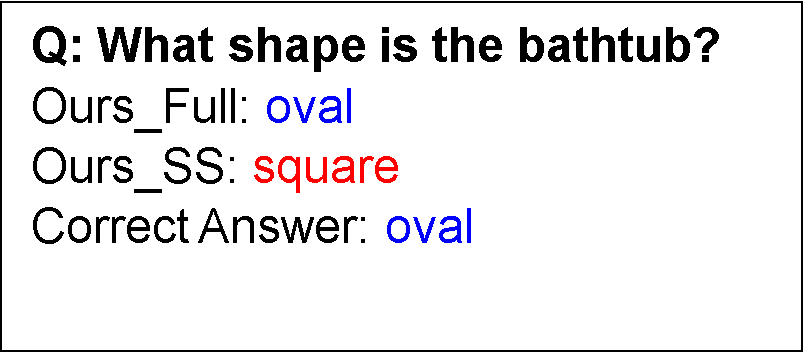} \\
\includegraphics[height=0.09\linewidth] {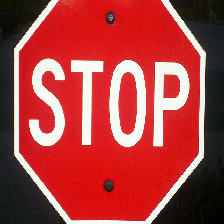}
\includegraphics[height=0.09\linewidth] {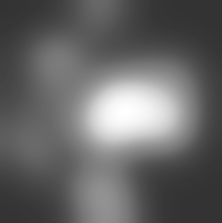}
\includegraphics[height=0.09\linewidth] {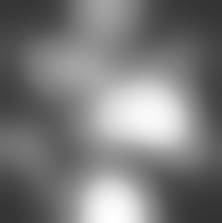}
\includegraphics[height=0.09\linewidth] {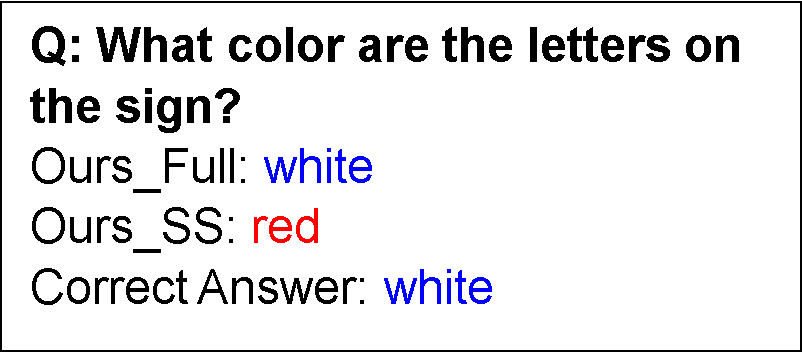} 
\includegraphics[height=0.09\linewidth] {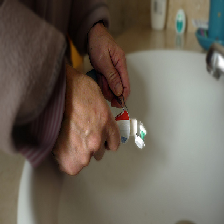}
\includegraphics[height=0.09\linewidth] {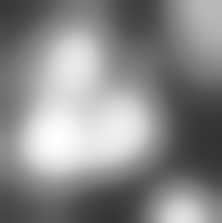}
\includegraphics[height=0.09\linewidth] {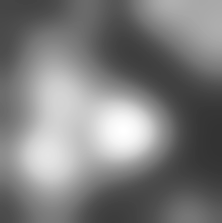}
\includegraphics[height=0.09\linewidth] {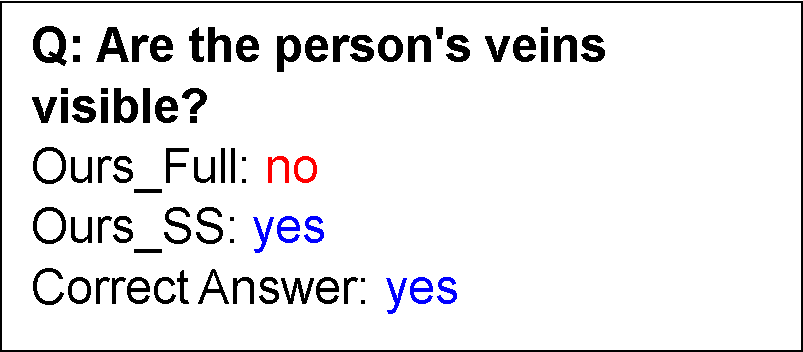} \\
\includegraphics[height=0.09\linewidth] {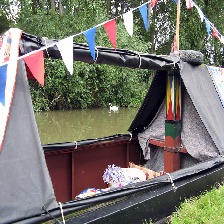}
\includegraphics[height=0.09\linewidth] {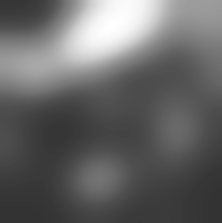}
\includegraphics[height=0.09\linewidth] {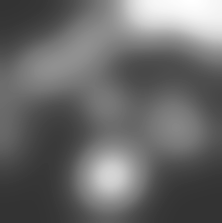}
\includegraphics[height=0.09\linewidth] {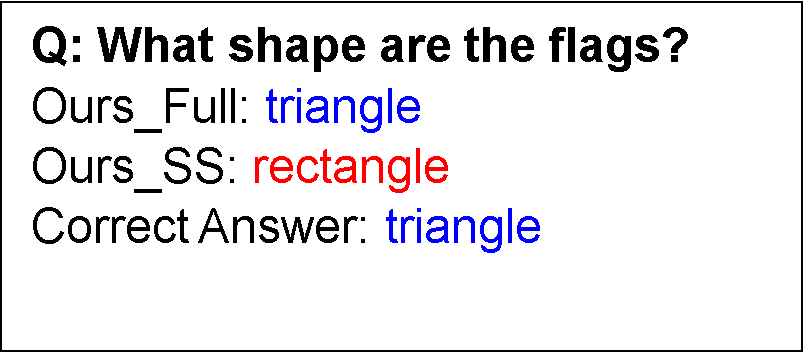} 
\includegraphics[height=0.09\linewidth] {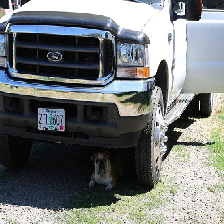}
\includegraphics[height=0.09\linewidth] {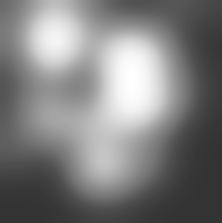}
\includegraphics[height=0.09\linewidth] {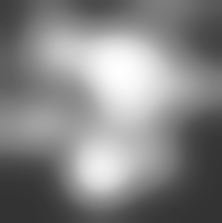}
\includegraphics[height=0.09\linewidth] {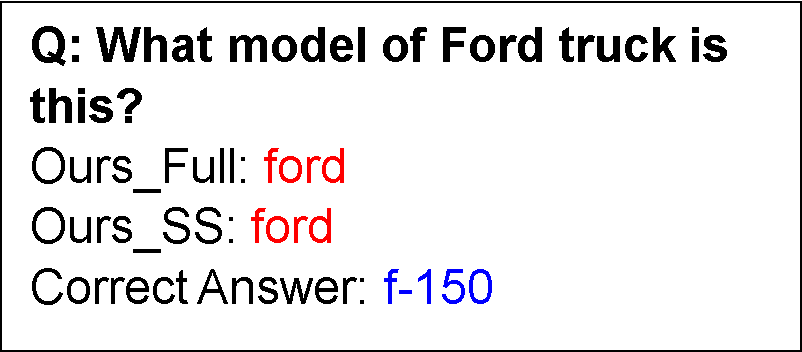} \\
\vspace{-0.2cm}
\caption{
Qualitative comparisons of attention between {\sf \small Ours\_Full} and {\sf \small Ours\_SS}.
}
\vspace{-0.3cm}
\label{fig:qualitative}
\end{figure*}

\end{multicols}

\end{document}